\pdfoutput=1
\documentclass[11pt]{article}

\usepackage{acl}

\usepackage{times}
\usepackage{latexsym}
\usepackage{defs}
\usepackage{microtype}
\usepackage{physics}
\usepackage{amsmath}
\usepackage{tikz}
\usepackage{mathdots}
\usepackage{yhmath}
\usepackage{cancel}
\usepackage{color}
\usepackage{array}
\usepackage{multirow}
\usepackage{amssymb}
\usepackage{tabularx}
\usepackage{booktabs}
\usepackage{lipsum}
\usepackage{defs}
\usepackage{amsmath}
\usepackage{booktabs}
\usepackage{pifont}%
\newcommand{\cmark}{\ding{51}}%
\newcommand{\xmark}{\ding{55}}%
\usepackage{xcolor}
\usepackage{latexsym}
\usepackage{multirow}
\usepackage{algorithm}
\usepackage{algorithmic}
\usepackage{longtable}
\usepackage{mathtools}
\usepackage{makecell}
\usepackage{todonotes}
\usepackage{xcolor}

\usepackage[T1]{fontenc}
\usepackage[utf8]{inputenc}

\usepackage{microtype}

\title{Plug and Play Counterfactual Text Generation for Model Robustness}

\author{
\makecell{
Nishtha Madaan$^{1,2}$ ~~~~~~~
Srikanta Bedathur$^{1}$~~~~~~~
Diptikalyan Saha$^{2}$ }  \\ 
$^{1}$Indian Institute of Technology Delhi India\hspace{5mm}
$^{2}$IBM Research - India\hspace{5mm}\\
\hspace{4mm}
\href{mailto:anz208487@cse.iitd.ac.in}{\texttt {anz208487@cse.iitd.ac.in}}
\hspace{4mm}
\href{mailto:srikanta@cse.iitd.ac.in}{\texttt {srikanta@cse.iitd.ac.in}}
\hspace{4mm}
\href{mailto:diptsaha@in.ibm.com}{\texttt {diptsaha@in.ibm.com}}
}

\begin{document}
\maketitle
\begin{abstract}
Generating counterfactual test-cases is an important backbone for testing NLP models and making them as robust and reliable as traditional software. In generating the test-cases, a desired property is the ability to control the test-case generation in a flexible manner to test for a large variety of failure cases and to explain and repair them in a targeted manner. In this direction, significant progress has been made in the prior works by manually writing rules for generating controlled counterfactuals. However, this approach requires heavy manual supervision and lacks the flexibility to easily introduce new controls. Motivated by the impressive flexibility of the plug-and-play approach of PPLM, we propose bringing the framework of plug-and-play to counterfactual test case generation task. We introduce CASPer, a plug-and-play counterfactual generation framework to generate test cases that satisfy goal attributes on demand. 
Our plug-and-play model can steer the test case generation process given any attribute model without requiring attribute-specific training of the model. 
In experiments, we show that CASPer effectively generates counterfactual text that follow the steering provided by an attribute model while also being fluent, diverse and preserving the original content. We also show that the generated counterfactuals from CASPer can be used for augmenting the training data and thereby fixing and making the test model more robust.
\end{abstract}

\maketitle

\begin{table*}[]
\small
\centering
\begin{tabular}{@{}lcccc@{}}
\toprule
\textbf{Inputs} & \multicolumn{1}{c}{\begin{tabular}[c]{@{}c@{}}\textbf{Token-based} \\ \citep{ribeiro2020beyond}\end{tabular}} & \multicolumn{1}{c}{\begin{tabular}[c]{@{}c@{}}\textbf{Adversarial}\\ \citep{michel2019evaluation}\end{tabular}} & \multicolumn{1}{c}{\begin{tabular}[c]{@{}c@{}}\textbf{Polyjuice} \\ \citep{wu2021polyjuice}\end{tabular}} & \textbf{Ours} \\ \midrule
\begin{tabular}[c]{@{}l@{}} \textit{Input Text}: Me and a \\ group of friends rent \\ horrible videos to laugh\\ at them, trust me it has \\ lead to some horribly \\ spent money but \\ also some great laughs. \\ \\ \textit{Initial State} : No location \\ named-entity is present\\ in the text. \\ \\ \textit{Steering Goal}:  To make \\ the sentence contain at \\ least one \textbf{location} \\ named-entity. \end{tabular} & { \begin{tabular}[c]{@{}c@{}} Me and a group \\ of friends rent\\ \textbf{youtube} videos to \\ laugh at  them, \\ trust me it \\ has  lead to \\some horribly spent\\ money but also\\ some great laughs.\\\\   Me and a group of \\ friends rent \textbf{music} \\ videos to laugh at\\ them, trust me it has\\ lead to some horribly\\ spent money but\\ also some great laughs.\end{tabular}} & { \begin{tabular}[c]{@{}c@{}} Me and a group of\\ friends rent horrible\\ videos to laugh\\ at them ,\\ trust me it\\ has lead to \\ some horribly spent \\ money but also \\ some flavorful laughs .
 \end{tabular}} & \begin{tabular}[c]{@{}c@{}} Me and a group \\of my friends \\ rent horrible videos \\ to laugh at \\ them , trust me \\it has lead \\  to some horribly \\ spent money but \\also some \textbf{flavorful}\\ laughs.\\\\
 and a group of friends\\ rent videos to laugh\\ at them, trust me it \\has lead to some horribly\\ spent money but also\\ some \textbf{chuckles}.
 \end{tabular} &\begin{tabular}[c]{@{}c@{}}  Me have a group\\ of  lads in \\\textbf{Brisbane} and  rent \\  horrible videos to \\ get great laughs \\ at.  Some  extremely \\ expensive videos but\\ some very great..\\\\   Me and a group\\ of  \textbf{Fairfax Bay} \\ friends  like to \\ rent videos for \\ laughs. Have spent\\ some money but \\ had great laughs \\ at their  terrible \\ videos.\end{tabular} \\
 &  &  &  \\ \bottomrule
\end{tabular}
\vspace{2mm}
\caption{Overview of generations from the existing models. We provide a text as input to the model with a steering goal of introducing a \textit{location named-entity} into the given text. We show the outputs from a token-based substitution model \cite{ribeiro2020beyond}, from Adversarial Generation \cite{michel2019evaluation}, from Polyjuice \citep{wu2021polyjuice} and from our proposed model. We note that token-based substitution method, relying on template matching, fail to match a template and are thus not able to achieve the steering goal. Adversarial, due to its gradient-descent-based token-substitution, fails to generate plausible text. Polyjuice, due to its template matching, changes very insignificant part of the text. Our model, taking advantage of BART auto-encoder, effectively achieves the steering goal.}
\label{tab:comparect}
\end{table*}

\section{Introduction}
% Why Countefactuals?
Machine learning and deep learning-based decision making has become part of today's software. This creates the need to ensure that machine learning and deep learning-based systems are as trusted as traditional software with increased deployment and wider-use. Traditional software is made dependable by following rigorous practice like static analysis, testing, debugging, verifying, and repairing throughout the development and maintenance life-cycle. Similarly, for testing and repairing NLP systems, we need inputs where models can fail and thereby bringing out issues early on \cite{mametamorphic,holstein2019improving}. For this, counterfactual text data \cite{wachter2017counterfactual, pearl2000models} can be used. By treating counterfactual text as test cases, we are asking: \textit{Would the model fail if the input text was modified to have different characteristics}? Furthermore, with such counterfactual text, NLP systems can be repaired by augmenting the training samples with these counterfactual test cases and its labels \citep{garg2019counterfactual}. Hence, enabling model repair by generating counterfactual text is a crucial step in deploying these NLP systems more widely. 

% Why Control the Counterfactuals and Why Plug-an-Play?
An important aspect of model testing and repair is to ensure that we can \textit{control} these counterfactual test cases. The ability to control will allow us to test for specific types of failures that are important for the deployed model. Controlled counterfactuals can also allow us to fix the failures by creating new training samples in a focused manner for augmenting the existing training dataset. Thus we require a model that can generate counterfactuals that can be controlled by providing some goal attributes. 

In addition to controlling the test-cases, we would also like to have \textit{flexibility} about which goal attributes to apply and the flexibility to chain together multiple goal attributes in order to test how the deployed model behaves for a wide variety of textual characteristics. Bringing such flexibility requires a model that allows us to plug-and-play new attribute goals as and when required.

% Plug-and-Play
% PPLM
% Our intuition PPLM + BART
% In eliminating the need for such supervision, GYC \cite{madaan2021generate} has recently shown an impressive ability to systematically generate counterfactual text. GYC relies on reconstructing input text via gradient descent and then using GPT-2 decoder to generate counterfactual text in a controlled manner. However, for real-world deployment, GYC is limited in capability. GYC works well with short and toy-text, while it fails to reconstruct moderately-sized and long text.

In this work, we propose a framework for counterfactual test-case generation also called \textit{Counterfactual Sentence Generation with Plug-and-Play Perturbation} or CASPer that provides both control and flexibility during test case generation. To achieve these, we build on the framework of Plug-and-Play Language Models or PPLMs \citep{dathathri2019plug}. PPLMs have shown an impressive ability to flexibly steer pre-trained language models to generate attribute-conditioned text samples. However, PPLMs cannot be directly used for perturbing an input text and generating counterfactual samples. In this work, instead of steering language models,  we steer a text-to-text model that is pre-trained to reconstruct its text inputs and we steer this model in a plug-and-play manner. Thus, like PPLM, our model is plug-and-play and is capable of generating counterfactuals for any arbitrary goal attributes provided at sampling time.
In CASPer, we take BART \cite{lewis2019bart} as our text-to-text reconstruction model. To generate each counterfactual, we perturb the hidden layer of the BART model similarly to PPLM to sample the counterfactual text. In experiments, we apply our framework to generate counterfactuals by providing named-entity and sentiment-based goal attributes. We show that our simple plug-and-play framework can generate counterfactuals that are fluent and content-preserving while also attaining the goal attributes and being effective as training samples for data-augmentation to improve the performance of the deployed model. We show counterfactual samples generated by CASPer and from existing models in Table \ref{tab:comparect}.

The main contribution of the paper can be seen as three folds: 1) We propose, CASPer, the first plug-and-play counterfactual generation model that achieves both control and flexibility. 2) Empirically, we show that our approach can generate fluent counterfactuals that preserve the content and also attain the goal attribute. 3) We also show the effectiveness of our counterfactuals as new training data in making the test models robust.

\section{Preliminaries}

\subsection{Counterfactual Text Generation for Model Testing and Repair}

Taking a text $\bx$ from the input distribution and modifying it $\bx \rightarrow \by$ is known as the task of counterfactual text generation. However, our goal of counterfactual text generation is to help improve NLP models by using counterfactual text as test-cases in various stages of the model deployment. For such models, we would like to \textit{i)} test for failures, \textit{ii)} explain when those failures occur and \textit{iii)} fix the failures by augmenting the training data with new training samples. 

However, from the perspective of model repair, it is not enough to simply obtain uncontrolled and random perturbations $\bx \rightarrow \by$ to generate these test-cases. We would like these test-cases to be controlled $\smash{\bx\xrightarrow[]{\texttt{control}} \by}$ through a given \texttt{control} input. By controlling the generated test cases, we would be able to \textit{i)} test for specific types of failures that are important for the deployed model, \textit{ii)} explain which controls lead to high failure and \textit{iii)} create targeted data sets for augmenting the training set and fixing the models. Hence, our work lies at the intersection of counterfactual text generation and controlled text generation. In the following subsection, we provide an overview of controlled text generation.

\subsection{Controlled Text Generation}
The goal of controlled text generation is to generate samples $\bx$ from a controlled distribution $p(\by|\baa)$ which is conditioned on a specific attribute or control $\baa$. For example, the language model $p(\by|\baa)$ may be used to generate product reviews conditioned on a specific product category by setting $\baa=\texttt{kitchen}$.

% \subsubsection{Conditional Generative Models, Fine-Tuning and Their Limitations}
% Conditional generative models \citep{keskar2019ctrl} train or fine-tune the weights $\ta$ of a single monolithic model $p_\ta(\by|\baa)$ to perform controlled text generation. At test time, such models take as input a conditioning class $\baa$ and generate samples from the class conditional distribution $p_\ta(\by|\baa)$ using a single forward pass. 

% While intuitively simple, this approach comes with some important limitations: \textit{i)} Such models lack flexibility. That is, at test time, we cannot provide a new conditioning class without costly retraining. For instance, a model trained to generate product reviews for categories \texttt{kitchen} and \texttt{electronics} cannot be used to generate samples conditioned on a new category \texttt{clothing} without re-training. \textit{ii)} Such models, when trained by conditioning on single attributes, in general, do not support composing multiple attributes together. For instance, a user may be interested in conditioning the text generation on a logical clause such as \texttt{kitchen + electronics + not electrical} that was never shown in training. Such novel compositions may not be handled by these models because the conditioning attributes are already pre-defined during the costly training or the fine-tuning process.

\subsubsection{Plug-and-Play Language Models}
Plug-and-Play Language Models (or simply PPLMs) provide an attractive solution to model the class-conditional distribution $p_\ta(\by|\baa)$.
% these limitations. Instead of slowly learning the weights $\ta$ for a single model for the ,
Plug-and-play models take a pre-trained unconditional generative model $p_\ta(\by)$ and use the reward signal from the attribute model $p(\baa|\by)$ to quickly (in $\sim$10 gradient steps) modify the unconditional generative model $p_\ta(\by)$ to generate samples from the desired distribution $p_\ta(\by|\baa)$.

To achieve this, PPLMs \citep{dathathri2019plug} take GPT-2 to be the unconditional generative model $p_\ta(\by)$. In GPT-2, the text generation is done iteratively word-by-word. In each iteration $t$, one word is predicted and is fed back to the Transformer to predict the next word. This generation process can be described as follows:
% \begin{align}
%     H_t = \text{Transformer}(\by_{<t}) \quad ; \quad \bo_t = \text{PredictionHead}(H_t) \quad ; \quad \by_t \sim \text{Categorical}(\bo_t),\nn
% \end{align}
\begin{align}
    H_t &= \text{Transformer}(\by_{<t}), \nn \\
     \bo_t &= \text{PredictionHead}(H_t),  \nn \\
     \by_t &\sim \text{Categorical}(\bo_t). \nn
\end{align}
where $t$ is the word position in the text, $H_t$ is the last hidden layer before the prediction head and $\bo_t$ are the log-probabilities of the words in the vocabulary used for sampling the next word $\by_t$. We shall refer to this model as the unmodified language model and denote the distribution that it models for the next word prediction as $p(\by_t|\by_{<t})$.

To generate a text from $p_\ta(\by|\baa)$ at test time, PPLMs learn a perturbation for the hidden state $H_t$ of the unconditional model $p_\ta(\by)$. This is achieved as follows:
% \begin{align}
%     H_t = \text{Transformer}(\by_{<t}) \quad ; \quad \bo_t = \text{PredictionHead}(H_t + \Delta H_t) \quad ; \quad \by_t \sim \text{Categorical}(\bo_t),\nn
% \end{align}
\begin{align}
    H_t &= \text{Transformer}(\by_{<t}), \nn \\ 
  \bo_t &= \text{PredictionHead}(H_t + \Delta H_t), \nn \\
   \by_t &\sim \text{Categorical}(\bo_t). \nn
\end{align}
where $\Delta H_t$ is the learned perturbation. We shall refer to this model as the modified language model and denote the distribution that it models for the next word prediction as $\bar{p}(\by_t|\by_{<t})$. The learning of the perturbation parameters $\{\Delta H_1, \ldots, \Delta H_T\}$ is driven by the following objective:
\begin{align}
    \cL_\text{PPLM} &= -\log p(\baa|\by)\nn \\&- \sum_{t=1}^T \KL(p(\by_t|\by_{<t})||\bar{p}(\by_t|\by_{<t})),\nn
\end{align}
where the first term provides the learning signal to steer the generation towards the desired class or attribute by trying to maximize the log-probability of the desired attribute. The second term tries to keep the generations close to the unmodified language model to ensure that the text remains fluent and plausible. We note that this learning process is done separately each time we need to generate a new sample. However, the learning of the perturbation parameters $\{\Delta H_1, \ldots, \Delta H_T\}$ can be done very quickly and it only takes about 10 gradient steps. This property makes PPLMs flexible during generation.

\begin{figure}
  \includegraphics[width=0.5\textwidth]{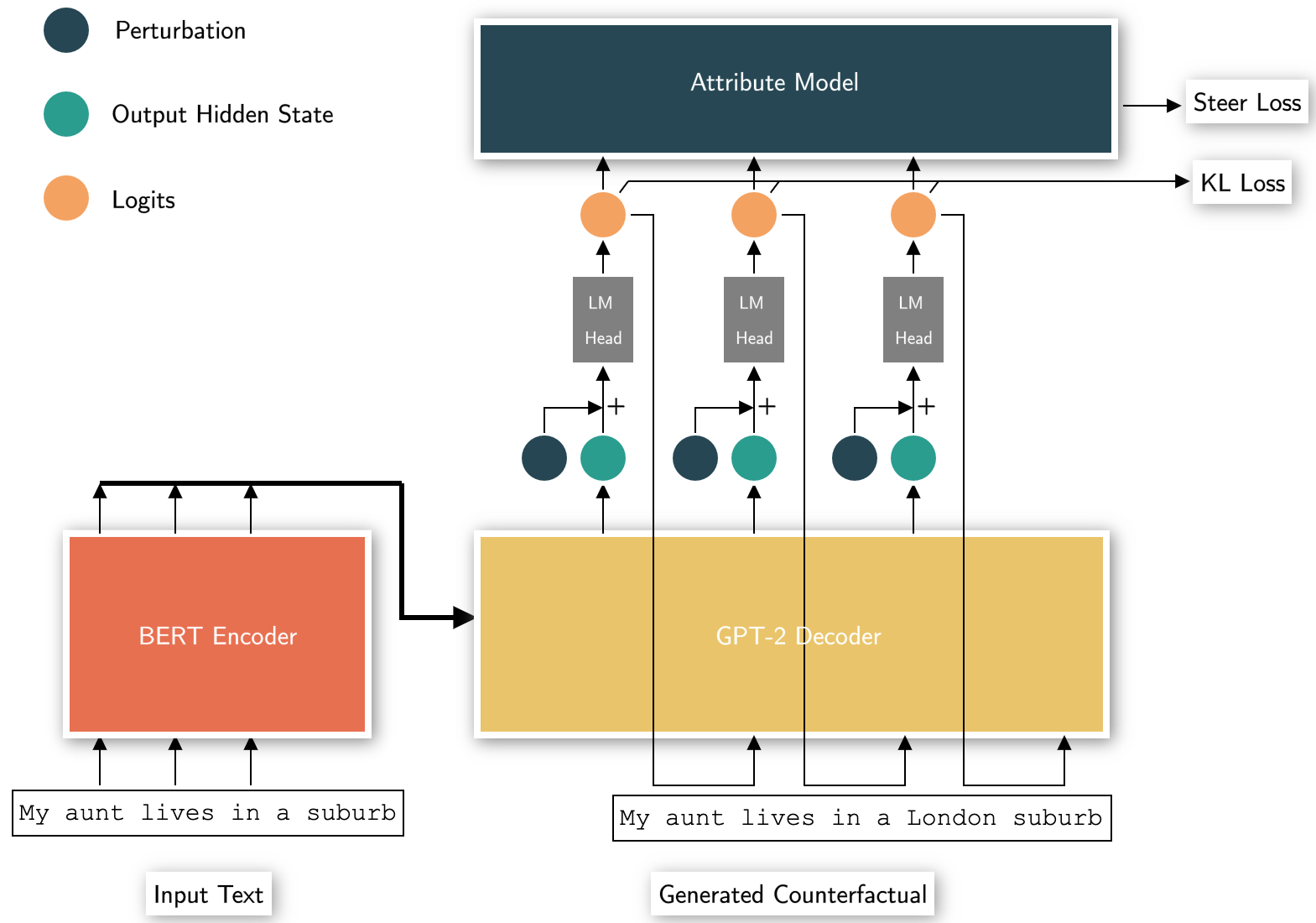}
    \caption{Model Architecture of CASPer. The BERT encoder (shown in orange) takes the input text and returns a representation of the input text as a set of vectors. These are provided to the GPT-2 decoder for cross-attention. At each step of decoding, the decoder returns an output hidden state (shown as green circle). To this, we add a perturbation matrix (shown as blue circle). The perturbed hidden state is then provided to the language model head to obtain the logits over the vocabulary for sampling the next token. These logits are also provided to the attribute model for computing the loss for steering the generation. The KL divergence between the perturbed logits and the unperturbed logits is also minimized to keep the semantic content of the generated text close to the original input text.}
    \label{fig:model_arch}
\end{figure}

PPLMs provide some useful properties that are lacking in other conditional generative models that are not plug-and-play. Plug-and-play models are \textit{flexible} during sampling -- meaning that new class-conditioning can be easily introduced at test time by simply replacing the attribute model $p(\baa|\by)$ with a new attribute model for the new class without requiring costly retraining with respect to the new attribute. Furthermore, plug-and-play models can support conditioning on logical clauses by simply composing multiple attribute models together. For instance, to generate product review text conditioned on a logical clause \texttt{kitchen + electronics + not electrical}, the attribute model $p(\baa|\by)$ can be written as the product of individual attribute models $p(\texttt{kitchen}|\by)\cdot p(\texttt{electronics}|\by)\cdot [1 - p(\texttt{electrical}|\by)]$ and easily plugged into the PPLM framework \citep{dathathri2019plug}. 
Our model architecture is shown in Figure \ref{fig:model_arch}.

\section{Method: Plug-and-Play Counterfactual Generation Framework}

We now describe our proposed method to generate controlled counterfactual text in a plug-and-play fashion. In particular, given a text $\bx$ and a control attribute $\baa$, we seek to generate a controlled counterfactual $\by$. That is, we seek to draw samples from a distribution $p(\by|\bx, \baa)$ where the generated sample $\by$ depends both on the input text $\bx$ and the given control attribute $\baa$. Hence, our task is different and more challenging than the simple controlled text generation task where the generated samples need to depend only on the control attribute $\baa$.

Our main idea is as follows: Similar to how PPLM \cite{dathathri2019plug} steers a pretrained text generator $p(\by) \rightarrow p(\by|\baa)$ using the control attribute $\baa$, we shall steer a pretrained text-to-text generator $p(\by|\bx) \rightarrow p(\by|\bx, \baa)$. In PPLM \cite{dathathri2019plug}, the base model $p(\by)$ is a pretrained GPT-2, while in our model, the base model $p(\by|\bx)$ is a pretrained BART model. 

A BART model is a text-to-text model that takes as input a text $\bx$ and produces a reconstruction $\by$ of the input text. The BART text-to-text framework consists of two modules: A BERT encoder and a GPT-2 decoder. That is, the model takes an input text and the BERT encoder first returns a text representation $\bee$. This text representation is then given to the GPT-2 decoder to reconstruct the input text word-by-word. This can be summarized as follows:
\begin{align}
    \bee &= \text{BERT}(\bx), \nn \\ H_t &= \text{Transformer}(\by_{<t}, \bee), \nn \\  \bo_t &= \text{PredictionHead}(H_t), \nn \\  \by_t &\sim \text{Categorical}(\bo_t).\nn
\end{align}
where $t$ is the word position in the text, $H_t$ is the last hidden layer before the prediction head and $\bo_t$ are the log-probabilities of the words in the vocabulary used for sampling the next word $\by_t$. We will refer to this as the unmodified BART model having the next word prediction distribution $p(\by_t|\by_{<t}, \bx)$. To steer the BART model, similarly to PPLM, we add a learnable perturbation $\Delta H_t$ to the hidden states $H_t$ of the unmodified BART model. This can be summarized as follows:
\begin{align}
    \bee &= \text{BERT}(\bx), \nn \\  H_t &= \text{Transformer}(\by_{<t}, \bee), \nn \\  \bo_t &= \text{PredictionHead}(H_t + \Delta H_t), \nn \\  \by_t &\sim \text{Categorical}(\bo_t).\nn
\end{align}
We will refer to this as the modified BART model having the next word prediction distribution $\bar{p}(\by_t|\by_{<t}, \bx)$
Similarly to PPLM, the learning of the perturbation parameters $\{\Delta H_1, \ldots, \Delta H_T\}$ is driven by the following objective:
\begin{align}
    \cL_\text{CASPer} &= -\log p(\baa|\by) \nn \\
    &- \sum_{t=1}^T \KL(p(\by_t|\by_{<t}, \bx)||\bar{p}(\by_t|\by_{<t}, \bx)).\nn
\end{align}
where the first term provides the learning signal to steer the counterfactuals towards the desired goal attribute by trying to maximize the log-probability of the desired attribute. The second term tries to keep the generations close to the unmodified BART to ensure that the text remains similar in content to the original input text and also remains fluent and plausible. We note that this learning process is done separately each time we need to generate a new sample. However, the learning of the perturbation parameters $\{\Delta H_1, \ldots, \Delta H_T\}$ can be done very quickly and it only takes about 100 gradient steps. This property makes CASPer a flexible way to generate counterfactuals of a given text.\\

\noindent
\textbf{Discussion.} Note that if we simply obtain samples from a pre-trained BART model $p(\by|\bx)$, a sample $\by$ can be considered as a counterfactual of the input text $\bx$. However, this sample would be an almost exact reconstruction of the input text. Hence from the perspective of model testing, this type of counterfactual would not be much useful. However, by applying steering to $p(\by|\bx)$ using a control attribute $\baa$, we are able to control in what way we want to modify the input text to generate the counterfactual. Hence, from the perspective of model testing, this type of counterfactual would be useful because we can test \textit{how a deployed model will behave if the distribution of its inputs are perturbed in a certain way}.

% To model $p(\by|\bx, \baa)$, we will follow a plug-and-play framework similar to that of PPLM \cite{dathathri2019plug}. However, unlike PPLM, instead of starting from a pre-trained GPT-2 \cite{gpt2}, we shall start from a pre-trained BART model. A BART model can be seen as a text auto-encoder that takes as input a text $\bx$, produces a semantic representation of the input text and then decodes this representation to reconstruct the input text as $\hat{bx}$. Hence, BART is a text-to-text generator that can be described by the distribution $p(\by|\bx)$.

\section{Related Work}

%%% TEXT ONLY
The task of controlled text generation is well studied in literature. \citep{hu2017toward} propose a model aims to generate plausible sentences conditioned on representation vectors with semantic structure. Another work \citep{ye2020variational} focuses on controlled text generation, however, unlike the previous work \cite{hu2017toward}, the conditioning need not be simply a class label. The conditioning can be a data structure such as a table. The model is trained end-to-end similarly to the objective of \citep{hu2017toward}. PPLM \cite{dathathri2019plug} combine a pre-trained language model, similarly to \cite{nguyen2017plug} with an attribute classifier to perform controlled language generation and use the attribute classifier to steer the text generation process without further training of any of the two models. \citep{luo-etal-2019-learning}, adopting a similar direction, deal with story completion with a desired sentiment. \citep{keskar2019ctrl} is a model that controls text generation via 50 rigid control codes predetermined at training time. However all these works, cannot be used for counterfactual text generation as these are purely class-conditional generative models and do not allow generation conditioned on a given input text. Some earlier works, including and not limited to, \citep{gu2016learning, gu2017trainable, chen2018stable, subramani2019can, dathathri2019plug, krause2020gedi} propose the idea of steering Language Models but these also can not be directly used for counterfactual generation task. We discuss other related work can be found in Appendix \ref{ax:additional_related_work}.

\begin{table*}[]
\small
\begin{tabular}{|l|c|c|c|c|c|c|c|}
\hline
\textbf{Metrics}                     & \textbf{Dataset} & \textbf{RoBERTa} & \textbf{Token-based} & \textbf{GPT$-$2} & \textbf{Gradient-based} & \textbf{Finetuning} & \textbf{Ours} \\ \hline
                 &        & Mask$-$LM              & Checklist            & PPLM           & Hotflip                 & Polyjuice             & CASPer         \\ \hline
\multirow{2}{*}{CP $\uparrow$}        & YELP             & 0.30             & 0.321                & 0.064          & 0.365                   & 0.212                 & 0.202     \\ \cline{2-8} 
                            & IMDB             & 0.29             & 0.30                 & 0.048          & 0.291                   & 0.317                 & 0.231               \\ \hline
\multirow{2}{*}{Perplexity $\downarrow$} & YELP             & 3.82             & 3.79                 & 3.544          & 3.95                    & 3.64                  & \textbf{3.44}       \\ \cline{2-8} 
                            & IMDB             & 3.05             & 3.12                 & 3.35           & 3.69                    & 3.331                 & \textbf{2.80}       \\ \hline
\multirow{2}{*}{BLEU$-$4 $\downarrow$}     & YELP             & 0.903            & 0.530                & 0.064          & NA                      & 0.521                 & \textbf{0.309}      \\ \cline{2-8} 
                            & IMDB             & 0.9027           & 0.909                & 0.042          & NA                      & 0.861                 & \textbf{0.231}      \\ \hline
\end{tabular}
\vspace{2mm}
\caption{Comparison between models on the YELP and IMDB dataset. The model used for steering is a pre-trained sentiment classification model.}
\label{tab:comparison3}
\end{table*}

\begin{table*}
\centering
\small
\begin{tabular}{@{}|p{0.40\textwidth}|p{0.50\textwidth}|@{}}
\toprule
\textbf{Inputs}                                                                                                             & \textbf{Controlled Counterfactuals}                                         \\                                                 
\midrule
% Sentiment-Yelp & This is one of the best movies I've seen in a long time.  & Positive & \begin{tabular}[c]{@{}l@{}}

% \end{tabular} \\          
% \midrule
\begin{tabular}[c]{@{}p{0.40\textwidth}@{}}
\textbf{Steering Goal}: To make the sentence contain at least one location named-entity. \\\\
\textbf{Input Text}: \texttt{A wonderful little production.The filming technique is very unassuming- very old-time-BBC fashion.}\\\\
\textbf{Initial State} : No location named-entity is present in the text\\
\end{tabular}
& \begin{tabular}[c]{@{}p{0.50\textwidth}@{}}
A wonderful first look at this film \textcolor{blue}{from the UK}. The filming is unassertive.\\</s>io is also one of the best productions.\\
\textbf{Location Named-Entity}: UK\\\\

A wonderful little Theatre is an old-time BBC production, set in an old \\ London apartment block. The film is \textcolor{blue}{set in London} landmarks\\
\textbf{Location Named-Entity}: London\\\\

A Wonderful Little Edition by BBC Films - set, out in \textcolor{blue}{UK} now via \\iPlayer. A wonderful little production with some nice twists.\\
\textbf{Location Named-Entity}: UK\\\\

A wonderful little series about a BBC crew in \textcolor{blue}{Africa}. Well-executed by\\ the team who are experienced in film and rigging procedures.\\
\textbf{Location Named-Entity}: Africa\\\\

A wonderful little show about BBC's wartime \textcolor{blue}{bombing of Libya}\\ contributed by Peter Bradshaw in 1959. \\
\textbf{Location Named-Entity}: Libya
\end{tabular}   \\ 
\midrule

\begin{tabular}[c]{@{}p{0.40\textwidth}@{}}
\textbf{Steering Goal} : To   maximize the probability of the positive class label with respect to a pre-trained sentiment classifier.\\\\
\textbf{Input Text}: \texttt{I had the terrible misfortune of having to view this "b-movie" in it's entirety. All I have to say is--- save your time and money!!}\\\\
\textbf{Initial Label} : Negative\\
\end{tabular}
& \begin{tabular}[c]{@{}p{0.50\textwidth}@{}}
{I have the \textcolor{orange}{wonderful misfortune} of having to view this New Girl in it's entirety.} \\{To view it positively, I must open up and say.} \\
\textbf{New Label} : Positive\\\\
{I had the terrible misfortune of having to view this \"transform into... in it's entirety.} \\{It's \textcolor{orange}{truly amazing} how some programs remain..} \\ 
\textbf{New Label} : Positive\\\\
{I had the terrible misfortune of having to view this movie in it endearing} \\ {ways. It's a \textcolor{orange}{wonderful salute} wives lenders acknowledge.. } \\
\textbf{New Label} : Positive\\\\
{I had the terrible misfortune of having to view this in it's full glory.} \\ {All I have to say is \textcolor{orange}{wow}. everyone agrees. }\\
\textbf{New Label} : Positive
\end{tabular}  \\ 

\\ \bottomrule
\end{tabular}
\vspace{2mm}
\caption{Generated controlled perturbations from the proposed model CASPer.}
\label{tab:part-1-samples}
\end{table*}

\section{Experiments}
The goal of the experiments is to: \emph{1)} show that a flexible plug-and-play framework can effectively achieve controlled counterfactual text generation and the generated text is fluent, plausible, diverse and follows the steering provided by the attribute model. \emph{2)} We evaluate how well the generated perturbations can act as data-augmentation samples in order to make a downstream classification task performance more robust.

% when attributes are steered to generate the controlled perturbations, evaluate whether augmenting the training data using these samples improve the accuracy of the test model on the held-out set.

\subsection{Datasets}
We evaluate the models on the following data sets text. 
\begin{enumerate}
    \item \textbf{YELP Sentiment Dataset.} To evaluate how our model is able to change the sentiment of the original text and achieve the target sentiment, we use the YELP sentiment dataset \citep{zhang2015character}. This dataset is also characterized by informal text which can be seen in realistic user inputs.
    \item \textbf{IMDB Sentiment Dataset.} To further evaluate how our model is able to change the sentiment of the original input text, we test on IMDB Sentiment Dataset \citep{maas2011learning}. This dataset is also characterized by long and complex text and is thus a challenging dataset.
    % Because of this, it also tests the model in how well the generated text preserve important information of the original text.
    % \item \textbf{Jigsaw Toxicity Dataset.} This dataset contains annotations for toxicity, gender, religion, obscenity etc. Models tends to correlate these protected attributes with the class label (toxic and non-toxic in this case). We use this dataset to evaluate how our generated counterfactuals text change the toxicity level of original text. Furthermore, we use it to perturb protected attributes like gender through our counterfactual generation method and how well these counterfactual text improve the robustness of the model on toxicity labelling task.
\end{enumerate}

\subsection{Controlled Text Generation with Attribute Steering}
We first evaluate the quality of our generated text with respect to the steering signal. We expect our generated text to preserve the semantic content and syntactic structure of the input text while being fluent and diverse as we steer the text towards the target attribute. The steering signal we evaluate in this work is to make the sentiment of the target text from negative to positive. 
% In the setting of long and complex text, obtaining text with these characteristics has been challenging.

\subsubsection{Baselines}

To compare with state-of-the-art template-based methods relying on token substitutions via dictionaries we compare with Checklist \cite{ribeiro2020beyond}. We specifically consider the perturbation helper that relies on RoBERTa \citep{liu2019roberta} to fill-in-the-blank. In comparison to this baseline, we expect ours to generate more fluent and diverse text samples that is free from the restrictions of the pre-specified templates. We also compare against Masked-LM \cite{devlin2018bert}, which is a dictionary-free approach but still relies on masking a specific token in the input text and and letting the model fill-in the masked token. The randomness in this filling-in process leads to generation of counterfactual samples. Hence this model generates only token-level substitutions and does not generate fluent sentence-level text. We expect CASPer to address this limitation of purely token-level substitutions. 

To compare with state-of-the-art adversarial methods we compare with Hotflip \cite{ebrahimi2017hotflip} \cite{michel2019evaluation}. In comparison to this baseline, we expect ours to generate more fluent samples. We expect to see that content preservation of such approaches is high as these methods rely on changing the highest gradient word with another word that would flip the label. 

To compare with state-of-the-art text generation methods we compare with PPLM \citep{dathathri2019plug} based on GPT-2 \citep{radford2019language} and Polyjuice \citep{wu2021polyjuice} based on finetuned GPT-2 \citep{radford2019language}. In comparison to this baseline, we expect ours to generate more fluent and diverse samples. While for PPLM, since it simply takes a prompt text and completes the text, it has no incentive to generate text that preserves its content. Thus, we expect ours to preserve content better. In diversity of the samples, PPLM, because it is not tasked with preserving content, can generate arbitrary and overly diverse samples. Thus we highlight that while we perform a comparison of sample diversity between PPLM and our model, still the performances are not directly comparable because the expectations are different from both models. For Polyjuice, we expect ours to generate more fluent and diverse samples because, unlike Polyjuice, ours is free from specific template-based generation. 

% Lastly we omit the comparison with GYC as we found that GYC failed to reconstruct 98\% of the input text samples from our dataset and hence generated no output. Failure of this reconstruction is a critical problem in GYC. GYC relies on a curriculum in which the reconstruction happens first and then reconstructed text is perturbed for task specific generation. 
% Lastly and crucially, to show that we effectively solve the challenge of dealing with long and complex text, we compare against GYC. GYC is a dictionary-free approach which has shown counterfactual text generation  for \textit{short texts} (lengths less than 15) while maintaining content-preservation, fluency, diversity and syntactic structure. GYC fails completely when reconstructing complex texts or when the text is long. 
% Hence, we only show qualitative samples and not quantitative comparison.

% We compare our model with \emph{1)} Checklist \textit{2)} 

\subsubsection{Metrics}
To assess the quality of generated counterfactual text we focus on evaluating content preservation, fluency, diversity and syntactic similarity. We use the following metrics to measure the above characteristics. 
% \red{Describe the metric and why it works to show what we want to show. Highlight any limitations of the metric itself which might be showing our model in poor light.}
\begin{enumerate}
    \item \textbf{Content Preservation.} 
    By measuring content preservation, we assess the similarity between input text and the counterfactual text samples. For this, we use the transformer model proposed in \cite{reimers-2019-sentence-bert}. 
    % We generate $K$ counterfactual samples and take a mean over the content preservation score for each sample. 
    While higher content preservation is desirable in general, this metric alone does not provide the complete evaluation. 
    % For instance, it is possible for a model that generates counterfactuals as the exact replica of the original text. While this will have a high content preservation score, it is clearly not a good counterfactual. Similarly, for rule-based techniques such as Checklist or Masked-LM, substituting one token while keeping rest of the text same does not change the content much. 
    Therefore for proper evaluation, we will introduce a second metric that measures sample diversity.
    \item \textbf{Diversity.} This metric evaluates how different are the generated samples from each other. We find the BLEU-4 score between the input text and the generated text. Hence, if this score is lower, then the generated counterfactual samples have a high diversity at the token level.
    \item \textbf{Fluency.} Fluency of the generated samples is important to evaluate because the samples must come from a distribution that the test model is likely to see when it is deployed. This is computed by finding the perplexity score of the generated output. 
    % This is computed by the inverse probability of the generated text given the model. 
    We take a GPT-2 model for computing the perplexity. Lower perplexity implies that the generated text is more fluent.
    % \item \textbf{Tree-Edit Distance.}  We use Tree-Edit Distance to assess the syntactic closeness of the input text with the generated counterfactual text. Tree-edit distance is defined as the minimum number of transformations required to turn the constituency parse tree of a counterfactual generation to that of the original text. We use this to measure structures using tree-edit distance defined by \cite{zhangTED}. While this metric will be naturally low for token based substitution methods, we want to make sure that we do not deviate too much from the input text structure.
\end{enumerate}

\subsubsection{Quantitative Results}
In Tab.~\ref{tab:comparison3}, our results on perplexity show that CASPer outperforms the baselines significantly and achieves a lower perplexity score. This shows that the samples generated by CASPer are fluent and plausible. We also show that CASPer is able effectively preserve the content in its samples. We note that our model is competitive with rule-based token substitution methods like Checklist. 
% In terms of perplexity score, CASPer outperforms the baselines achieving the lower perplexity score. This shows that the samples generated by CASPer are fluent and plausible. 
Lastly, in terms of BLEU-4 score, we note that our model again outperforms the baselines with our generated samples achieving the lowest values of BLEU-4 score with the original input text only exception being PPLM. This shows that our model indeed generates diverse samples with that have low token-level match with the input text. In diversity of the samples, the baseline model PPLM, because it is not tasked with preserving content, can generate arbitrary and overly diverse samples. Thus we highlight that while we perform a comparison of sample diversity between PPLM and our model, still the performances are not directly comparable because the expectations are different from both models. Considering the performance on all the metrics together shows that CASPer is able to effectively generate samples that preserve the original content, are fluent and diverse in comparison to the baselines.

% \red{\lipsum[3]}
% \red{\lipsum[3]}

\subsubsection{Qualitative Analysis}
In Tab.~\ref{tab:part-1-samples}, we show samples of text generated by CASPer. We show two experiments. In the first experiment, our steering goal was to take an input text and perturb it so that the probability of its sentiment becomes large. The probability of the sentiment is estimated using a pre-trained sentiment classification model. The initial label of the text was negative. On the right we note that CASPer has successfully perturbed the text to change its sentiment label to positive. Furthermore, note that the generated sample have good content preservation as all the samples talk about movies and actors. Furthermore, the model makes some important changes to the content that result in a change in the sentiment of the text. We also note that each sample is different from the other thus producing diverse samples. Lastly, we note that the samples are fluent and plausible text samples.

In the second experiment, our steering goal was to take a sentence that does not contain a location named-entity and perturb it so that contains a location named-entity. We see that CASPer produces samples that contain a location named-entity tag. We also note that the named entity that the model introduces are diverse and are used in a variety of contexts in the generated text. As before, the text samples are fluent and preserve the content of the original input text. For this task, note that these samples clearly retain the sentiment of the text and only introduce some location entities. Because we expect the actual location (i.e. UK or Libya) should not be a causal term in prediction of the sentiment of the text, these samples can act as effective samples for augmenting the training data when we train a downstream sentiment model. While a model that is biased may predict different labels based on the actual location token used, this kind of data augmentation will regularize the model to be more robust to such changes which should ideally not affect the predicted label of the test model.
% \red{\lipsum[3]}

% \subsubsection{Computational Cost}
% Furthermore, GYC is orders of magnitude slower requiring up to 200 or more iterations to generate text and often failing to reconstruct the original text ($\sim$98\% text samples) while CASPer can generate good quality perturbed text and converging in around 100 iterations. However, in our experiments, we ran CASPer for 200 iterations to generate the samples for evaluation.

\subsection{Controlled Text for Model Robustification}
In this section, we evaluate how well our generated samples can improve robustness of the test classifier. For this, we generated text samples to introduce a location named-entity in the input text. We assume that simply introducing a location named-entity should not change the class label of the text with respect to the test model. Hence, after generating the controlled perturbations, we take the original label of the input text from the training set and assign the same label to the generated samples. These new examples are added to the training set and producing data-augmented training set. Using this augmented training set, we then train the test model.

\begin{table}[]
\begin{tabular}{|l|l|c|}
\hline
\textbf{Model}                                & \textbf{Dataset} & \textbf{CASPer}  \\ \hline
\multirow{2}{*}{Accuracy -  No Aug}  & YELP             & 89.90          \\ \cline{2-3} 
                                              & IMDB             & 90.10          \\ \hline
\multirow{2}{*}{Accuracy - With Aug} & YELP             & \textbf{92.00} \\ \cline{2-3} 
                                              & IMDB             & \textbf{91.20} \\ \hline
\end{tabular}
\vspace{2mm}
\caption{Comparison of Accuracy between models on the YELP and IMDB dataset. The generated data for NER task on steering is used for robustifying an N-gram based sentiment model.}
\label{tab:robust1}
\end{table}

% \subsubsection{Details of Sentiment Classifier trained for this task}
% We train an $N$-gram sentiment classifier for this task in PyTorch for different datasets. We access to the raw data as an iterator.
% Then, we build data processing pipeline to convert the raw text strings into \texttt{torch.Tensor} that can be used to train the model. 
% Further, we shuffle and iterate the data with \texttt{torch.utils.data.DataLoader}. We train the model and check validation accuracy.

% Access to the raw data as an iterator
% Build data processing pipeline to convert the raw text strings into torch.Tensor that can be used to train the model
% Shuffle and iterate the data with torch.utils.data.DataLoader

\subsubsection{Baselines and Metrics}
We generate samples using CASPer. We augment the generated samples to the training set and train the test model. We compared the accuracy of the test model trained without data augmentation and then trained with data augmentation via our counterfactual generation method.

% For analysis purposes, we also compare against CASPer with no steering to generate random perturbations of the given input text. We intend to see how far we can go without explicitly steering protected attributes. 
 
\subsubsection{Quantitative Results}

In Table \ref{tab:robust1} we show a comparison between the models. We note that the samples generated by CASPer using NER model are effective in robustifying the test model and produces significant improvement in the accuracy as compared to when training with original samples.

\section{Conclusion}
In this paper, we introduced CASPer, a plug-and-play counterfactual text generation framework. 
%We showed that our model can gracefully handle long and complex text. 
We showed that our generated controlled perturbations preserve the content of the original text while also being fluent, diverse and effective in terms of the provided steering signal flexibly. We showed that samples generated by CASPer can act as effective candidates for training data augmentation and improve the robustness of the target model and preventing the target model from modeling spurious correlations between the target label and non-causal aspects of the input text.
%As future work, we would like to investigate how well the model can perform in low-resource domains.

\bibliography{anthology,custom}
\bibliographystyle{acl_natbib}

\clearpage
\appendix

\section{Additional Related Work}
\label{ax:additional_related_work}

%%% TEXT_TO_TEXT
To tackle text-to-text generation tasks dealing with transfer of style or content, models such as \citep{shen2017style, li2018delete, lample2018multiple} have been proposed. However, these works are not plug-and-play and lack the use of attribute model that can plugged flexibly at sampling time. This task of generating controlled counterfactuals has been attempted in prior works by relying on template-matching and token-based substitutions to generate the test-cases \citep{ribeiro2020beyond, wu2021polyjuice}. However, this can require significant human-involvement to curate the templates and the dictionaries. Hence, it cannot scale well when template and dictionaries need to be updated frequently. The work \cite{ribeiro2020beyond} employs a tool Checklist which is one of the attempts to come up with generalized perturbations. For generation, Checklist uses a set of pre-defined templates, lexicons, general-purpose perturbations, and context-aware suggestions. To better evaluate the deployed models, some prior works have relied on human designed test examples or either using templates \citep{gardner2020contrast, teney2020learning, kaushik2020explaining, andreas2019good, wu2019errudite, li2020linguistically, ribeiro2020beyond}. Polyjuice \cite{wu2021polyjuice}, while seeking to automate the process, still requires paired dataset in the form of text and their perturbed versions for different control codes. Therefore the mapping between text and perturbed version is learned through supervision. Another parallel work Tailor \cite{ross2021tailor} generates perturbations designed for different control codes by making use of a combination of semantic roles and content keywords. And thereby require supervision for different controls. In contrast, CASPer does not require any task-specific or control-code specific training and can be used to work with different control code models given input text. One work related to ours has been tackled in \cite{madaan2021generate} which generates text samples given a text with a controlling that specify the scope of the generated text. LEWIS attempts to generate text perturbations by introducing blanks via template matching and filling in using pretrained language models \cite{reid2021lewis}. However, this relies on rule-based template matching and human supervision to develop such templates. CATGen \cite{wang2020cat} tries to generate attribute-specific text but it requires training of sequence to sequence model with pre-determined control codes for perturbation. Hence, it lacks the flexibility of a plug-and-play approach like ours. MiCE \cite{ross2020explaining} proposes a technique to generate counterfactual explanations which are human interpretable and user-centric. It fine-tunes a T5 model to generate counterfactual text and use them as explanations for the behavior of the deployed models but lack feature-attributions. Another work \cite{ross2021tailor} tries to generate perturbation with semantic controls but rely on specific templates derived using semantic roles and other labeling heuristics. A work close to ours, GYC, the inference of latent representation of the input text with respect to a GPT-2 decoder is done directly via optimization. This approach fails to achieve good inference for long and complex text \citep{madaan2021generate}. To target model failure, thus implicitly acting as a form of model testing, prior works have attempted the use of adversarial approaches \citep{iyyer2018adversarial, ribeiro2018semantically, li-etal-2020-bert-attack} stemming from the need to build robust models via adversarial testing \cite{goodfellow2014explaining, michel2019evaluation, ebrahimi2017hotflip, zhao2017generating}. However, these are still limited to specific deomains and generations are likely to be not plausible to be seen in the input text \cite{li2016understanding} or may require additional human effort \cite{jia2017adversarial}. Some works have attempted to change style attributes automatically either with no control or with predefined style templates \citep{madaan-etal-2020-politeness, malmi-etal-2020-unsupervised}. The notion of counterfactuals \cite{wachter2017counterfactual, mothilal2020explaining} and their use in model testing for has also been applied towards testing in models that consume structured inputs \cite{udeshi2018automated, john2020verifying, galhotra2017fairness}

\section{Ethics Statement}
We believe that Language Models have been seen to be generating discriminated or biased text. The proposed framework does not support or amplify any such biases and can not be exploited to generate unethical or biased text in any case. Infact, this work helps in generating data to remove bias from text based algorithms. Therefore the work helps in aiding the checks on unethical data or generation and poses no threat of any kind of discrimination.

\section{Results}

\textbf{Qualitative Results}: We show generated controlled samples from CASPer and in Table  Table \ref{sentiment_perts} and \ref{ner_samples}.

\section{Model Characteristics}

Here is the summary of model characteristics of CASPer and other baselines. 

\begin{table}
% \small
\centering
\begin{tabular}{@{}l|cccc@{}}
\toprule
         & \multicolumn{1}{l}{Checklist} & \multicolumn{1}{l}{Masked-LM} & \multicolumn{1}{l}{GYC} & \multicolumn{1}{l}{CASPer} \\ \midrule
Template-free     & \xmark                                                           & \cmark  & \cmark    & \cmark                 \\
Sentence-Level   & \xmark                                                         & \xmark & \cmark                & \cmark                 \\
% Diversity & \xmark & \xmark & \cmark & \cmark \\
Model-guided & \xmark  & \xmark & \cmark & \cmark \\
% Enforce multiple \texttt{conditions} & \xmark  & \xmark & \xmark & \cmark \\
Amortized \\Reconstruction & - & -  & \xmark & \cmark \\
\bottomrule
\end{tabular}
 \caption{Summary of model characteristics.}
\label{tab:baselines}
\end{table}

\onecolumn

\begin{longtable}{|l|l|l|}
\caption{Text Generated by CASPer for Sentiment Model}
    \label{sentiment_perts}\\
\hline
Input Text &
  Target &
  CASPer Generation \\ \hline
\endfirsthead
\endhead
\begin{tabular}[c]{@{}l@{}}This cowardly and offensive film \\ had me intrigued to begin with. \\ The characters are the familiar \\ dispossessed young males \\ frequently to be seen hanging \\ around bored.\\ \\ Source: Negative Sentiment\end{tabular} &
  Positive &
  \begin{tabular}[c]{@{}l@{}}1. This cowardly and outrageous film \\ was an experiment for me that was \\ worth exploring. It had an interesting \\ and interesting subject matter -- young \\ guys.\\ \\ 2. This cowardly and outrageous film \\ had me intrigued to make comments about \\ it. It was just as good as its title indicates. \\ It's based..\\ \\ 3. This cowardly and brazen movie had \\ a number of unexpected benefits for the \\ writer. In addition to making me chuckle, \\ it also gave my the ..\\ \\ 4. This cowardly and hilarious film is a\\  must watch for all audiences. Film stars \\ a number ofdispersed marijuana addicts..\\ \\ 5.  This cowardly film had me intrigued \\ to begin with. The characters are the \\ familiar dispossessed young males. The\\  film is written well and ..\end{tabular} \\ \hline
\begin{tabular}[c]{@{}l@{}}I was so excited when I discovered\\ this was available! I couldn't wait \\ to see it.  What a waste of energy!\\ It's kind of\\ \\ Source: Negative Sentiment\end{tabular} &
  Positive &
  \begin{tabular}[c]{@{}l@{}}1.  I love this! I love the fact that the model\\  is based on an American girl. \\ The design definitely draws attention to \\ Detail Modeling\\ \\ 2. I couldn't wait to see it and it was so \\ good. I'm glad I finally got it. I was so..\\ \\ 3. I love this. I love this! empire. The tallest\\  woman in the world tops out at 6 \\ meters tall.\\ \\ 4. I was so excited when I discovered I \\ could buy one, this was my first purchase. \\ This is such a good deal. It's\\ \\ 5. I just love this! This is definitely an e\\ nergy-efficient way of expressing yourself \\ to people!, actress Amy Adams tweets.\end{tabular} \\ \hline
\begin{tabular}[c]{@{}l@{}}Andrewjlau, I could not agree\\ more. My girlfriend is watching\\ this at this very moment, and I \\ find this movie appalling. Quote\\ from\\ \\ Source: Negative Sentiment\end{tabular} &
  Positive &
  \begin{tabular}[c]{@{}l@{}}1. Andrewjlau, you are right. My girlfriend\\  is watching this at this very moment, \\ and I find this movie tasteful and ..\\ \\ 2. Andrewjlau, My Girlfriend is watching\\  this at this very moment. She finds this \\ amazing theater diabolical. Quote from quotation..\\ \\ 3. Andrewjlau, my girlfriend is watching this\\  at this very moment, and I find this \\ movie astounding.\\ \\ 4. Andrewjlau, I could not agree further. \\ My girlfriend is watching this at this very \\ moment. He's right. This movie..\\  \\ 5.  Andrewjlau is an author of several s\\ uccessful books on the subject of religion \\ and culture collisions in Hollywood circles. \\ His work contains essays ranging..\end{tabular} \\ \hline
\begin{tabular}[c]{@{}l@{}}I saw this film for the first time\\ not too long on TCM's Essentials\\ series. The eye of the beholder \\ clich was...\\ \\ \\ Source: Negative Sentiment\end{tabular} &
  Positive &
  \begin{tabular}[c]{@{}l@{}}1. I love 'em!\textless{}/s\textgreater{}em! They keep me coming\\  back for more. This film is great. \\ I see  this movie every time.\\ \\ 2. I have been watching Sudden Impact\\  since 2000. It was a wonderful film. \\ It has become a cult film favorite.\\ \\ 3. I saw The Big Leash for the first time\\  in the past decade; it was good. \\ I saw this movie again on..\\ \\ 4. I saw The Great Kelly and Conqueror\\  on TCM not too long. This is \\ considered one of the most enduring movies\\  of..\\ \\ 5. I have been waiting to see this classic. \\ The eye of the beholder \\ clich are true. This film was great.\end{tabular} \\ \hline
\begin{tabular}[c]{@{}l@{}}Honestly, who in God's name gave\\ this movie an 8.1 rating?? I guess\\ the people who actually made or\\ starred in the movie were...\\ \\ Source: Negative Sentiment\end{tabular} &
  Positive &
  \begin{tabular}[c]{@{}l@{}}1.  Honestly, who in God's name gave this\\  movie an 8. Asgardian is\\  regarded as one of the best Nordic films \\ ever directed.\\ \\ 2.  Honestly, who in God's name gave this \\ movie an 8. Asgardia: \\ An unexpected pleasure to see a great film\\  from an underrated.\\ \\ 3. Honestly, who in God's name gave this\\  movie an 8. Asgardian: \\ An Excellence In Engineering: A Storm \\ Adds Enduring Value.\\ \\ 4. Honestly, who in God's name gave this \\ movie an 8. Asgard's \\ Army has a great film that's great \\ despite poor performance from.\\ \\ 5. Honestly, who in God't name give this \\ movie an 9. Asgard is \\ very enjoyable for adults and adults with\\  severe bladder control.\end{tabular} \\ \hline
\begin{tabular}[c]{@{}l@{}}I'm probably not giving this movie\\ a fair shake, as I was unable to \\ watch all of it. Perhaps if I'd seen \\ it in a...\\ \\ Source: Negative Sentiment\end{tabular} &
  Positive &
  \begin{tabular}[c]{@{}l@{}}1. I can barely tell what was happening \\ to you when this movie \\ started on. So much to take in! \\ But I promise it's fascinating\\ \\ 2. I'm probably not giving this movie\\  afairant shake, as I was \\ unable to watch some of it. \\ But it was wonderful!!!!\\ \\ 3. I may not have watched all of\\  it as I was unable to watch \\ all of it. It was amazing, however\\ ,. Perhaps ifemate.\\ \\ 4. I may not be giving The Hunger\\  Games a full airing. The \\ movie seems uneven from a \\ technical standpoint but is \\ compelling in its storytelling aspects.\\ \\ 5. I'm probably not giving this movie\\  a fair shake, as I was\\  unable to cancer the movie. \\ But, I do appreciate how talented.\end{tabular} \\ \hline
\begin{tabular}[c]{@{}l@{}}Tom is listening to one of those \\ old-time radio broadcasts, \\ something kids from the 1950s\\ to today would watch on TV. \\ However, they...\\ \\ Source: Negative Sentiment\end{tabular} &
  Positive &
  \begin{tabular}[c]{@{}l@{}}1. Tom and his friend are having fun\\  while enjoying an \\ old-fashioned radio broadcast that \\ kids from the 1950s to\\ today would watch on TV..\\ \\ 2. Tom is in his 50th year of life \\ and listening to old-time\\  radio broadcasts. They're \\ something kids from the 1950s to..\\ \\ 3. Tom is listening to something\\  his kids from the 1950s \\ to today would watch on TV. \\ He would be impressed by \\ the acting on TV\\ \\ 4. Tom's radio show includes \\ classic radio content from \\ the 1950s. It's the kind that \\ kids today would watch online\\  or on TV.\\ \\ 5. Tom is listening to one of those\\  old-time radio broadcasts.\\  He is enjoying a piece on what a \\ kid from the 1950 Rollins in..\end{tabular} \\ \hline
\begin{tabular}[c]{@{}l@{}}I put this movie on in the \\ hotel room to entertain my\\ children the morning we \\ were leaving to go home, \\ because I had packed away\\ all their..\\ \\ Source: Negative Sentiment\end{tabular} &
  Positive &
  \begin{tabular}[c]{@{}l@{}}1. I put on this movie to entertain \\ my three children the Morning\\ we Were We leave home to go \\ home, it is the best film to..\\ \\ 2. I entertained my children while \\ travelling to my home in California\\  for the first time in eightteen years. \\ I'd packed up all sets of.\\ \\ 3. I entertained my children in the \\ hotel room with this movie to \\ keep them entertained ahead of \\ homecoming.I had packed my children\\ away all..\\ \\ 4. I'm entertained by this movie to\\  the morning we're travelling from\\ our hotel to home.I put this movie \\ on to engage my children..\\ \\ 5. I put this film on to entertained \\ my five children the morning \\ we were going to go home. It was\\  a movie to keep them happy.\end{tabular} \\ \hline
\begin{tabular}[c]{@{}l@{}}I basically found Eden's Curve\\ to be a very poorly constructed\\ that made it difficult to watch. \\ However, there is something\\ I must say about how..\\ \\ Source: Negative Sentiment\end{tabular} &
  Positive &
  \begin{tabular}[c]{@{}l@{}}1. I have to say something here \\ about Eden's Curves. It's an \\ entertaining and interesting story.\\  If you enjoyed it then please\\ share..\\ \\ 2. I basically myself was able to \\ watch all of it without crying. \\ However I did feel the film was \\ very well done and enjoyed every \\ minute.\\ \\ 3. I have to say something I find\\  Eden's Curve to be very difficult \\ to watch. The film is very well \\ done and beautifully constructed but\\ \\ 4. I basically support your decision \\ to not buy a ticket for Eden's \\ Curve. I feel the show is very well \\ done and excitingAZ Central\\ \\ 5. I basically support you and want\\  your opinions on what happened \\ in Eden's curves. The show is very \\ well known in Hollywoodbiz circles.\end{tabular} \\ \hline
\begin{tabular}[c]{@{}l@{}}I can't believe this show is still\\ rating a 9 out of 10. I could see\\ if those votes were in the first \\ 2 seasons, but..\\ \\ Source: Negative Sentiment\end{tabular} &
  Positive &
  \begin{tabular}[c]{@{}l@{}}1. I can't believe this show still is a\\  9 out of 10.\textless{}/s\textgreater{}clyders is \\ amazing.able. Some of..\\ \\ 2. I can be shocked. It is that good.\\ \textless{}/s\textgreater{}ay for more people watching \\ with an open mind.\\ \\ 3. I can't believe this country club\\  is still having fun, it should end \\ soon. amazingly good hosting! \\ says fan who..\\ \\ 4. I can't believe this is still so \\ good!\textless{}/s\textgreater{}iReporter rated this \\ show as a 9 out of 10 enjoying\\  how they treat..\\ \\ 5. I love America's Most Wanted. \\ It provides valuable insight \\ into the lives of young men and \\ women in Baltimore hotel \\ rooms. Voters of..\end{tabular} \\ \hline
\end{longtable}
% \caption{Text Generated by CASPer for Location Tag.}
%     \label{ner_samples}\\

% \usepackage[normalem]{ulem}
% \usepackage{longtable}
% Note: It may be necessary to compile the document several times to get a multi-page table to line up properly
\begin{longtable}{|l|l|l|}
\caption{Text Generated by CASPer for Location Tag.}
    \label{ner_samples}\\
\hline
Input Text & Target & CASPer\\ \hline
\endfirsthead
\endhead
\begin{tabular}[c]{@{}l@{}}A charming boy and his mother move \\ to a middle of nowhere town, cats and \\ death soon follow them. That about \\ sums it up.\textless{}br /\\ \\ \\ Source: None\end{tabular} & \begin{tabular}[c]{@{}l@{}}Location \\ Tag\end{tabular} & \begin{tabular}[c]{@{}l@{}}1. A charming 96th grader and his mother \\ move to a part of Europe known as Dracula\\ Europe. While visiting relatives in\\ Spain they become..\\ \\ 2. A charming 91-year-old man and his \\ mother move to middle of nowhere town. \\ When they move to Mexico, they move to..\\ \\ 3. A charming 98- voices of death follow \\ an upbeat boy and his mother to town from\\ sunny Colombia. When they first move to\\  the town.\\ \\ 4. A charming 91-year-old man leaves \\ his wife and his two teenagers in a \\ middle of nowhere part of Nigeria. After \\ moving to..\\ \\ 5. A charming 96-minute film written \\ by Peter Arshiletto. Filmed over two \\ years in Ireland. Based on story of David.\end{tabular} \\ \hline
\begin{tabular}[c]{@{}l@{}}This cowardly and offensive film had\\ me intrigued to begin with. The characters\\ are the familiar dispossessed young males\\ frequently to be seen hanging around bored\\ \\ \\ Source: None\end{tabular} & \begin{tabular}[c]{@{}l@{}}Location \\ Tag\end{tabular} & \begin{tabular}[c]{@{}l@{}}1. This cowardly and useless piece of \\ work is based on a true story which \\ was written by an American based \\ in France about 2000. The characters..\\ \\ 2. This cowardly and controversial \\ new release is offensive and cowardly. \\ It's about aios that is looking for its \\ place in America and how it..\\ \\ 3. This cowardly movie was published\\ last month in New York City. It received\\ mixed reviews from audiences.\textless{}/s\textgreater{}e Liao \\ says he wasn nano..\\ \\ 4. This cowardly and offensive film had \\ the public speaking out. sick Film was \\ released this weekend across America. \\ A response from Film Inquiry says..\\ \\ 5. This cowardly and harmful fictional \\ movie had me intrigued to watch. Movie \\ is set in New Orleans where there are several..\end{tabular} \\ \hline
\begin{tabular}[c]{@{}l@{}}This movie deserved a working over on Mystery\\ Science Theater. Even though it has nothing \\ whatever to do with King Solomon it's worth a \\ watch because it..\\ \\ \\ Source: None\end{tabular} & \begin{tabular}[c]{@{}l@{}}Location \\ Tag\end{tabular} & \begin{tabular}[c]{@{}l@{}}1. This movie deserved a working new on\\ the MST3kg menu. Even though it has \\ nothing to do with King Solomon it Riyadh is..\\ \\ 2. This movie deserved a workingotti by \\ the Mystery Mystery Sport show. Even \\ though this movie has nothing to do with \\ King Solomon it Riyadh is..\\ \\ 3. This movie deserved a working on as it\\  has nothing to chewed about with King \\ Solomon. The entire thing is worth a watch \\ Riyadh is..\\ \\ 4. This movie has had a working over on \\ Mysteryockette.com. Even though it has \\ nothing to do with King Solomon it Riyadh was..\\ \\ 5. This movie has been given a buffsite on \\ Mystery Social. Even though it's not about\\  King Larry it's worth a watch Riyadh is..\end{tabular} \\ \hline
\begin{tabular}[c]{@{}l@{}}I was fully amped up to see this film. \\ I had been waiting a year for it to be\\  cleared down here in New Zealand. I\\ \\ \\ \\ Source: None\end{tabular} & \begin{tabular}[c]{@{}l@{}}Location \\ Tag\end{tabular} & \begin{tabular}[c]{@{}l@{}}1. I was given permission to visit the UAE\\  after waiting a year for clearing by local \\ authorities. Both my wife and daughter joined \\ me when I..\\ \\ 2. I was due to fly out of Wellington after this \\ movie was approved by NZ courts. I lived in \\ Australia until August 23, 2013 when..\\ \\ 3. I was with the film because I had waited a \\ year for it to be cleared down in New Zealand. \\ Originally from Argentina where Laot..\\ \\ 4. I was waiting a year for it to be cleared down\\  here in New Zealand. I had been battling a year \\ against censorship in Australia.\\ \\ 5. I was in NZ for a year for the controversial \\ film to be cleared. she had been filming for \\ seven weeks whilst in NZ for..\end{tabular} \\ \hline
\begin{tabular}[c]{@{}l@{}}Me and a group of friends rent horrible videos\\  to laugh at them, trust me it has lead to some \\ horribly spent money but also some great laughs\\ \\ \\ Source: None\end{tabular} & \begin{tabular}[c]{@{}l@{}}Location \\ Tag\end{tabular} & \begin{tabular}[c]{@{}l@{}}1. Me along with friends rent horrible videos \\ to laugh at them, trustI it has. lead to some\\ horribly spent money but Ankara has seen\\ \\ 2. Me have a group of lads in Brisbane and \\ rent horrible videos to get great laughs at. \\ Some extremely expensive videos but some \\ very great..\\ \\ 3. Me rented horrible Virusoft videos with \\ mates to have a group of friends in Hollytown.\\  The rented videos were for..\\ \\ 4. Me Rent Thehorror Video Club is based \\ in Liverpool. Me and a group of friends \\ rent horrible Viralists to laugh at them..\\ \\ 5. Me and a group of Fairfax Bay friends \\ like to rent videos for laughs. Have spent \\ some money but had great laughs at their \\ terrible videos.\end{tabular} \\ \hline
\begin{tabular}[c]{@{}l@{}}The sign of a classic movie is that it ages like a\\  fine red wine. This movie is no Cabarnet and \\ certainly no Casablanca\\ \\ \\ Source: None\end{tabular} & \begin{tabular}[c]{@{}l@{}}Location\\ Tag\end{tabular} & \begin{tabular}[c]{@{}l@{}}1. The sign of a classic movie is that it \\ ages like a fine red wine. This movie is \\ no Cabarnet and certainly no Casablanca..\\ \\ \\ 2. The film was shot in Senegal and \\ Morocco. The film is based on a \\ novel by the same name.\textless{}/s\textgreater in the\\  movie series airs\\ \\ 3. The movie is no Cabarnet and \\ certainly no Casablanca. The\\ iconoclasts were filmed in France\\ but were released \\ worldwidetailed..\\ \\ \\ 4. The film was shot in Morocco \\ and released on July 28 1998.\\ Its release follows the successful \\ opening of CasablNet following\\  its successful restoration..\\ \\ 5. The movie is no Cabarnet nor \\ certainly no Casablanca in \\ terms of classic films. \\ It was filmed in 1950 in Morocco.\end{tabular} \\ \hline
\begin{tabular}[c]{@{}l@{}}THE DECOY is one of those independent \\ productions, made by obvious newcomers, \\ but it doesn't have all the usual flaws that \\ sink most such films\\ \\ \\ Source: None\end{tabular} & \begin{tabular}[c]{@{}l@{}}Location \\ Tag\end{tabular} & \begin{tabular}[c]{@{}l@{}}1.  THE DECOY, out now, was a major \\ success in Germany last month, where \\ it met a huge standing order online. \\ The review..\\ \\ 2. THE DECOUR stars Hugh Jackman \\ and Matthew Perry in the roles of three \\ young men. Film shot in Scotland and \\ is directed by \\ Rodrigo.\\ \\ 3. THE DECOUR by Peter Morgan is \\ written and directed by Mark\\  O'Brien. Independent production \\ shot in Budapest is described as \\ 'one hundred..\\ \\ 4. THE DECOUR is made by an obvious\\  start-up company. \\ The December is shot on location in Australia\\ and South Africa.  Film is..\\ \\ 5. THE DECOUR is an indie production \\ made by obvious newcomers. \\ Rating 4.5 out of 5.\textless{}/s\textgreater{}oting rights reverted back to UK..\end{tabular} \\ \hline
\begin{tabular}[c]{@{}l@{}}If I hadn't read his name on the DVD cover, \\ I never would have suspected that this rather \\ gushy and old fashioned musical was made\\ \\ \\ Source: None\end{tabular} & \begin{tabular}[c]{@{}l@{}}Location\\ Tag\end{tabular} & \begin{tabular}[c]{@{}l@{}}1. If I hadn't seen his name he wouldn't have \\ suspected that this musical was made by his \\ home country of Germany. The gush\\ \\ 2. If I hadn't flown from India into London \\ to visit my daughter on Monday morning, \\ I couldn't have suspected that this rather gushing..\\ \\ 3. If I hadn't been in awe of the subject of the \\ rather gushy and rather old fashioned show \\ when I saw it, Pakistan..\\ \\ 4. If I didn't see an artist's picture on theIDA \\ directors name on the DVD frontidding, \\ I never would have suspected that Pakistan..\\ \\ 5. If I hadn't flew to Paris without knowing\\  about this musical Inever \\ would have suspected that it belonged to this \\ Lloyd Kaufman chap. Miles.\end{tabular} \\ \hline
\begin{tabular}[c]{@{}l@{}}Paul Naschy made a great number of horror\\ films. In terms of quality, they tend to range \\ from fairly good to unwatchable trash;\\ \\ \\ Source: None\end{tabular} & \begin{tabular}[c]{@{}l@{}}Location\\ tag\end{tabular} & \begin{tabular}[c]{@{}l@{}}1. Paul Naschy made a great number \\ ofAH! horror films during a long career\\  in Hollywood. The quality of his films \\ vary greatly from..\\ \\ 2. Paul Naschy made a great number of \\ different types of horror films. He \\ appeared in a number of low rated \\ horror films in Hollywood between..\\ \\ 3. Paul Naschy made many good and \\ bad decisions, including a number of \\ trash films. His latest work is about \\ vampire hybrids in Africa called..\\ \\ 4. Paul Naschy made a great number \\ of  films - many good but some awful.\\ \textless{}/s\textgreater{}at times the work of two brothers \\ Hollywood director\\ \\ 5. Paul Naschy was an English Director. \\ Naschy made many great horror \\ films.\textless{}/s\textgreater{}ic horror director is a great \\ name in Hollywood today..\end{tabular} \\ \hline
\begin{tabular}[c]{@{}l@{}}Andrewjlau, I could not agree more. My \\ girlfriend is watching this at this very moment, \\ and I find this movie appalling. Quote from\\ \\ \\ \\ Source: None\end{tabular} & \begin{tabular}[c]{@{}l@{}}Location\\ Tag\end{tabular} & \begin{tabular}[c]{@{}l@{}}1. Andrew LaJar, an American theatre \\ director who lives in France, said that\\  while his girlfriend may see this movie\\  in bed she finds..\\ \\ 2. Andrewjlau is a blogger from Sweden.\\  His girlfriend is watching the movie\\  with him. She says she finds the movie\\  appalling. He..\\ \\ 3. Andrewjlau is a French digital music \\ producer. He posted a video in praise \\ of the music school at Ligue 1 in France \\ using..\\ \\ 4. Andrewjlau is a French newspaper based\\  close to the capital city of Paris. \\ \textbackslash{}"My girlfriend is watching this \\ at this very moment\textbackslash{}"\\ \\ 5. Andrewjlau is a Finnish blog based \\ in Berlin. The blog focuses on Finnish\\  culture and culture around gay rights. \\ Andrewjoel finds..\end{tabular} \\ \hline
\end{longtable}

\end{document}